\documentclass{article}

\usepackage{arxiv}

\usepackage{cite}

\newcommand{\citep}[1]{\cite{#1}}
\newcommand{\citet}[1]{\cite{#1}}

\usepackage{amsmath,amssymb,amsfonts}
\usepackage{graphicx}
\usepackage{textcomp}

\usepackage{dsfont}
\usepackage{nicefrac}

\usepackage{booktabs}
\usepackage{float}
\usepackage{multirow}
\usepackage{array}
\usepackage{xcolor}
\usepackage{csquotes}

\usepackage{algorithm}
\usepackage[noend]{algpseudocode}

\algrenewcommand\alglinenumber[1]{\scriptsize #1:}

\usepackage{url}
\usepackage{tikz}
\usetikzlibrary{calc, positioning, arrows.meta}

\newcommand{\githubrepo}{\url{https://github.com/jakobbossek/dynvrp}}

\title{Towards Decision Support in Dynamic Bi-Objective Vehicle Routing}
\date{ }

\author{
   Jakob Bossek \\
  Optimisation and Logistics\\
  The University of Adelaide\\
  Adelaide, Australia \\
  \texttt{jakob.bossek@adelaide.edu.au} \\
   \And
   Christian Grimme \\
   Information Systems and Statistics \\
   University of M\"unster \\
   M\"unster, Germany \\
   \texttt{christian.grimme@wi.uni-muenster.de}
   \And
   G\"unter Rudolph \\
   Dept. of Computer Science \\
   TU Dortmund University \\
   Dortmund, Germany \\
   \texttt{guenter.rudolph@tu-dortmund.de}
   \And
   Heike Trautmann \\
   Information Systems and Statistics \\
   University of M\"unster \\
   M\"unster, Germany \\
   \texttt{heike.trautmann@wi.uni-muenster.de}
}

\begin{document}
\maketitle

\pagestyle{plain}
\thispagestyle{fancy}
\lfoot{\vspace*{-1.25cm}\rule{\columnwidth}{0.2pt}\\\footnotesize \textcopyright2020 IEEE. Personal use of this material is permitted. Permission from IEEE must be obtained for all other uses, in any current or future media, including reprinting/republishing this material for advertising or promotional purposes, creating new collective works, for resale or redistribution to servers or lists, or reuse of any copyrighted component of this work in other works.\\ This version has been accepted for publication at the \textit{IEEE Congress on Evolutionary Computation (IEEE CEC)} 2020, which is part of the \textit{IEEE World Congress on Computational Intelligence (IEEE WCCI)} 2020.}\cfoot{}

\begin{abstract}
We consider a dynamic bi-objective vehicle routing problem, where a subset of customers ask for service over time. Therein, the distance traveled by a single vehicle and the number of unserved dynamic requests is minimized by a dynamic evolutionary multi-objective algorithm (DEMOA), which operates on discrete time windows (eras). A decision is made at each era by a decision-maker, thus any decision depends on irreversible decisions made in foregoing eras. To understand effects of sequences of decision-making and interactions/dependencies between decisions made, we conduct a series of experiments. More precisely, we fix a set of decision-maker preferences $D$ and the number of eras $n_t$ and analyze all $|D|^{n_t}$ combinations of decision-maker options. We find that for random uniform instances (a) the final selected solutions mainly depend on the final decision and not on the decision history, (b) solutions are quite robust with respect to the number of unvisited dynamic customers, and (c) solutions of the dynamic approach can even dominate solutions obtained by a clairvoyant EMOA. In contrast, for instances with clustered customers, we observe a strong dependency on decision-making history as well as more variance in solution diversity.
\end{abstract}

\keywords{Transportation \and vehicle routing \and decision making \and multi-objective optimization \and combinatorial optimization \and orienteering \and dynamic optimization}

\section{Introduction}
\label{sec:introduction}

In industry, especially logistics \citep{GFK17} and 24/7 production \citep{MPGL06}, many decisions must be made under uncertain knowledge about future events. Production planning problems may include tasks appearing over time, which is often expressed via release dates~\citep{pinedo2012scheduling}, while logistic problems like the traveling salesperson problem (TSP) may be augmented with dynamically changing traffic~\citep{Cheong2012} or -- more abstract -- with dynamically changing distance matrices~\citep{YKCL2004}.

Here, we consider the scenario that repeated decision requests are not independent from the previous ones; rather, every decision reduces the number of possible actions/decisions in future affecting the quality of the best possible solution
that can be achieved.

If optimization methods are used to support the decision process, an optimizer could iterate for short time intervals (eras). After each era, a human \emph{decision maker} (DM) may decide whether the proposed solution is executed or not. 
This situation is studied in the field of \emph{interactive optimization}~\citep{MKFPG15} (headword: human-in-the-loop).

\newpage
In analyzing the quality of optimization algorithms it is quite common to replace the human DM by a software agent (automatic decision maker)~\citep{BOGNMM18} to speed up experiments or to make it amenable to a theoretical analysis.
The software agent may obey some simple guidelines or even a complex set of user-supplied preferences to arrive at a decision.
The situation becomes even more complex in case of multiple objectives~\citep{BDMS08}; if the multi-objective optimizer uses the a-posteriori approach, the (automatic) DM must additionally decide, which non-dominated solution should be picked from the Pareto-front.

Here, we also take the approach of replacing the human by an automatic decision maker in case of a dynamic bi-objective vehicle routing problem, where the goal is to minimize both the distance traveled by a single vehicle and at the same time minimize the number of unvisited customers which ask for services over time. This special variant of the TSP problem includes the additional problem of subset selection of serviced customers and is similar to the so-called TSP with profits~\citep{Aghabeig2018}. However, the problem considered here comprises a dynamically growing set of customers, who request service over time.

It is important to note, that the focus of this work is not on the performance of the optimization algorithm but (1) on the impacts of specific (automatic) decision making rules on the final solution and (2) the visualization of subsequent decisions and solutions as a preliminary step towards an (interactive) decision support system. The dynamically generated solutions are compared to the solution of the so-called clairvoyant optimizer which has complete knowledge about the future, i.e., request times of dynamic customers.
In our study we run experiments for all possible combinations of a fixed set of automatic DM decision rules to scrutinize as many aspects of the decision rule as possible.

The
work is organized as follows: Section \ref{sec:probForm} details the dynamic multi-objective problem before describing the dynamic multi-objective evolutionary optimization algorithm supporting the decision maker in Section~\ref{sec:MOEA}.
The experimental setup (including the automatic decision maker) and results are described in Section~\ref{sec:experiments}.
The conclusion and the prospects of this work towards inclusion in interactive decision support systems are presented in Section~\ref{sec:conclusion}.

\section{Problem Description}
\label{sec:probForm}

We consider a dynamic vehicle routing problem for which the overall goal is to have a single service vehicle visiting customer locations from the set $C$ of all customer locations.

The dynamic character of this problem originates in the time-dependent appearance of customers from $C$. The set of customers $C\setminus\{1,N\} = C^m \cup C^o$ is divided into two disjoint subsets: Mandatory customers $C^m$ are known at time $t=0$ and must be visited by the vehicle while dynamic customers $c \in C^o$ ask for service at request time $r_t(c), t > 0$ as time passes by. They can either be visited or not.
In a real-world context, we may imagine the vehicle as a customer service vehicle with fixed orders and spare time to handle dynamically emerging service requests.

We assume that the vehicle leaves w.l.o.g. at a start depot $1 \in C$ and ends at an end depot $N \in C$.\footnote{This assumption is more general than starting and ending at the same depot, although a circular tour may be the normal case in real-world scenarios.}
The optimization task is to (1) minimize the overall tour length and at the same time (2) minimize the number of unserved dynamic customers.
Undoubtedly, the goals conflict with each other and we are faced with a complex dynamic combinatorial multi-objective optimization problem (MOP), for which we strive to find a set of (near) optimal compromises. Here, we adopt the notion of \emph{Pareto-dominance} for a definition of optimal compromise solutions: for two vehicle tours $x$ and $y$ we say that $x$ dominates $y$, if $x$ is not worse in any objective and strictly better in at least one objective~\citep{Coello2006}. The set of all non-dominated solutions is termed the \emph{Pareto-set}, its image in the objective space is called the \emph{Pareto-front}. Hence, in our scenario for each point in time $t \geq 0$ a bi-objective problem needs to be solved and the problem can be fully described by the sequence of all trade-off solution sets. Since time is continuous, this approach is infeasible in practice. A common approach is to discretize the time horizon, i.~e., the time interval in which dynamic customers pose requests, into a number of phases $n_t$, so-called \emph{eras}, of length $\Delta \in \mathbb{R}_{\ge 0}$ (see, e.~g. \cite{RY13}).

At the beginning of each era $j$ time $t = (j-1)\cdot \Delta$ has already passed and we may consider the set $C^m \cup C^o_{\leq t}$, with $C^o_{\leq t}$ being the set of dynamic customers, which asked for service before time $t$, as a static MOP. Tackling this static MOP with the algorithm of our choice results in an approximation of the Pareto-set. Finally, a decision maker (DM) is given the resulting set of trade-off solutions in each era $j$ and has to decide on how to guide the vehicle on the road until the beginning of the next era where then, more knowledge about dynamic customers becomes available. A crucial aspect here is that in each era $j > 0$ time already passed and consequently the vehicle might already have visited a subset of mandatory and/or optional customer locations. These decisions are irreversible and (1) may have a strong impact on the achievable solution quality (this is because a part of the solution space may become infeasible) and (2) exhibit a strong dependence on the decisions made by the DM in foregoing eras.

Static formulations of bi-objective vehicle routing problems or so-called traveling salesperson problems with profits~\cite{Feillet2005} have been addressed by several authors so far, e.~g. proposing exact $\varepsilon$-constraint methods~\cite{BGP09}, approximations schemes~\cite{FS13} or meta-heuristics~\cite{JGL08,Aghabeig2018}. Additionally, dynamic decision making gained some attention in the context of vehicle routing problems in general~\cite{Pillac2013,Meisel2011}. However, work on the intersection, i.~e. dynamic multi-objective vehicle routing problems is still rare. Braekers et al.~\cite{Braekers2016} show in their extensive literature review that less then 3\% of the literature between 2009 and 2015 address dynamic aspects. According to them, mentionable research includes work by \citet{Wen2010}, \citet{Lorini2011}, \citet{Khouadjia2012}, \citet{Hong2012}, and~\citet{Barkaoui:2013}. This list may be extended by mostly evolutionary approaches of~\citet{NH18}, \citet{Ghan2014}. Although of dynamic nature, the problem formulations have very diverse characteristics like moving service time windows, multiple vehicles, or changing structures of the network.
Own work addressed a clairvoyant and non-dynamic variant of the here discussed problem with an evolutionary multi-objective algorithm (EMOA)~\cite{MGB2015}. We enhanced the EMOA in a follow-up work by local search integration into the evolutionary search process~\citet{BGMRT2018}. This clairvoyant approach is considered here as reference approach. A sophisticated dynamic variant was presented in \cite{BGMRT2019}.


\section{The Dynamic Multi-Objective Evolutionary Algorithm}
\label{sec:MOEA}
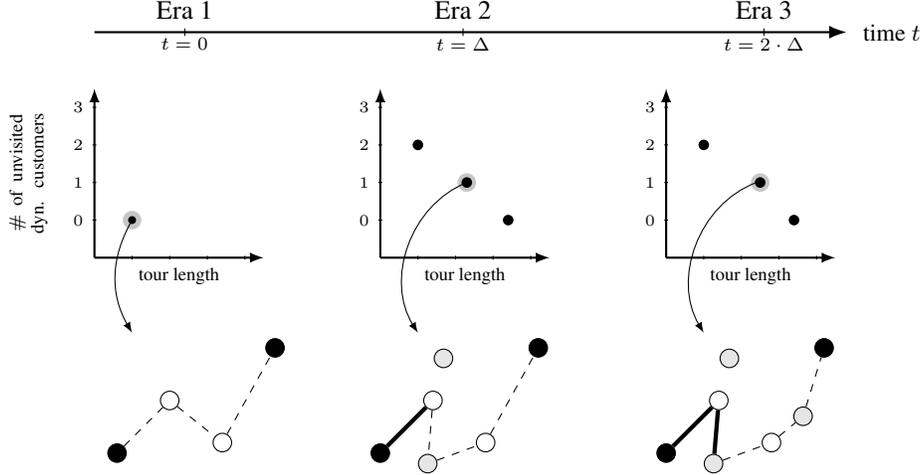
\begin{figure}[tb]
  \centering
  \begin{tikzpicture}[scale=1]
    \begin{scope}[shift={(0,0.3)}]
      \draw (0, 0) edge[-Latex, thick] (10, 0);
      \node at (10.6, 0) {\footnotesize{time $t$}};

      \draw (1.2, 0.05) edge[-] (1.2, -0.05) node[below] {\scriptsize{$t = 0$}} node[above] {Era 1};

      \draw (4.9, 0.05) edge[-] (4.9, -0.05) node[below] {\scriptsize{$t = \Delta$}} node[above] {Era 2};

      \draw (8.9, 0.05) edge[-] (8.9, -0.05) node[below] {\scriptsize{$t = 2 \cdot \Delta$}} node[above] {Era 3};
    \end{scope}

    \begin{scope}[shift={(0, -2.7)}, scale=0.5]
      \draw[thick, -latex] (0, 0) -- node[below] {\scriptsize{tour length}} (4.5, 0);
      \draw[thick, -latex] (0, 0) -- (0, 4.5);
      \node[rotate=90, text width=2.5cm, text centered] at (-1.75, 2.25) {\baselineskip=9pt\scriptsize{$\#$ of unvisited dyn. customers}\par};

      \fill[gray!45, dashed] (1, 1) circle(7pt);
      \fill (1, 1) circle(3pt) node[above] {};

      \foreach \i/\j in {0/1, 1/2, 2/3, 3/4}
      {
        \draw (\j, 0.05) -- (\j, -0.05);
        \draw (0.05, \j) -- (-0.05, \j) node[left] {\tiny{$\i$}};
      };

      \draw (1, 1) edge[-latex, bend right=30] (1, -2);
    \end{scope}

    \begin{scope}[shift={(3.8, -2.7)}, scale=0.5]
      \draw[thick, -latex] (0, 0) -- node[below] {\scriptsize{tour length}} (4.5, 0);
      \draw[thick, -latex] (0, 0) -- node[above,rotate=90, yshift=10pt] {} (0, 4.5);

      \fill (1, 3) circle(4pt) node[above] {}; 
      \fill[gray!45, dashed] (2.3, 2) circle(7pt);
      \fill (2.3, 2) circle(4pt) node[above] {}; 
      \fill (3.4, 1) circle(4pt) node[above] {}; 

      \foreach \i/\j in {0/1, 1/2, 2/3, 3/4}
      {
        \draw (\j, 0.05) -- (\j, -0.05);
        \draw (0.05, \j) -- (-0.05, \j) node[left] {\tiny{$\i$}};
      };

      \draw (2.3, 2) edge[-latex, bend right=50] (1, -2);

    \end{scope}

    \begin{scope}[shift={(7.6, -2.7)}, scale=0.5]
      \draw[thick, -latex] (0, 0) -- node[below] {\scriptsize{tour length}} (4.5, 0);
      \draw[thick, -latex] (0, 0) -- node[above,rotate=90, yshift=10pt] {} (0, 4.5);

      \fill (1, 3) circle(4pt) node[above] {}; 
      \fill[gray!45, dashed] (2.5, 2) circle(7pt);
      \fill (2.5, 2) circle(4pt) node[above] {}; 
      \fill (3.4, 1) circle(4pt) node[above] {}; 

      \foreach \i/\j in {0/1, 1/2, 2/3, 3/4}
      {
        \draw (\j, 0.05) -- (\j, -0.05);
        \draw (0.05, \j) -- (-0.05, \j) node[left] {\tiny{$\i$}};
      };

      \draw (2.4, 2) edge[-latex, bend right=50] (1, -2);

    \end{scope}

    \begin{scope}[every node/.style={circle, draw, font=\scriptsize, inner sep=1pt,minimum size=7pt}, shift={(0.3,-5.3)}, scale=.7]

      \node[fill=black] (DEP1) at (0, 0) {};
      \node[fill=black] (DEP2) at (3, 2) {};

      \node[fill=white] (M1) at (1, 1) {};
      \node[fill=white] (M2) at (2, 0.2) {};

      \draw[dashed] (DEP1) -- (M1) -- (M2) -- (DEP2);
    \end{scope}

    \begin{scope}[every node/.style={circle, draw, font=\scriptsize, inner sep=1pt,minimum size=7pt}, shift={(3.8,-5.3)}, scale=.7]

      \node[fill=black] (DEP1) at (0, 0) {};
      \node[fill=black] (DEP2) at (3, 2) {};

      \node[fill=white] (M1) at (1, 1) {};
      \node[fill=white] (M2) at (2, 0.2) {};

      \node[fill=gray!20] (D1) at (0.9, -0.2) {};
      \node[fill=gray!20] (D2) at (1.2, 1.8) {};

      \draw[dashed] (DEP1) -- (M1) -- (D1) -- (M2) -- (DEP2);
      \draw[ultra thick] (DEP1) -- (M1);
    \end{scope}

    \begin{scope}[every node/.style={circle, draw, font=\scriptsize, inner sep=1pt,minimum size=7pt}, shift={(7.6,-5.3)}, scale=.7]

      \node[fill=black] (DEP1) at (0, 0) {};
      \node[fill=black] (DEP2) at (3, 2) {};

      \node[fill=white] (M1) at (1, 1) {};
      \node[fill=white] (M2) at (2, 0.2) {};

      \node[fill=gray!20] (D1) at (0.9, -0.2) {};
      \node[fill=gray!20] (D2) at (1.2, 1.8) {};
      \node[fill=gray!20] (D3) at (2.6, 0.7) {};

      \draw[dashed] (DEP1) -- (M1) -- (D1) -- (M2) -- (D3) -- (DEP2);
      \draw[ultra thick] (DEP1) -- (M1) -- (D1);
    \end{scope}

  \end{tikzpicture}

  \newcommand{\dmchoice}{%
    \begin{tikzpicture}[scale=0.5, baseline=-3mm, thick]
      \fill[gray!45, dashed] (0, 0) circle(4pt);
      \fill (0, 0) circle(2pt);
    \end{tikzpicture}%
  }

  \caption{Exemplary illustration of three eras of the DEMOA. The scatterplots show the Pareto-front approximations at eras $j = 1, 2, 3$. Here, solutions selected by the DM are highlighted (\raisebox{-2pt}{\protect\tikz{\protect\fill[gray!45] (0, 0) circle(4pt); \protect\fill (0, 0) circle(2pt);}}).
  Below, the selected tours are visualized (depots \raisebox{-1pt}{\protect\tikz{\protect\fill (0,0) circle(3pt);}}, mandatory customers \raisebox{-1pt}{\protect\tikz{\protect\draw[fill=white] circle(3pt);}} and dynamic customers \raisebox{-1pt}{\protect\tikz{\protect\draw[fill=gray!45] circle(3pt);}}). The thick solid path highlights the fixed partial tour already driven by the vehicle at the beginning of the corresponding era. Note that in this example solutions with three unvisited customers are infeasible in the 3rd era since one dynamic customer is already served.}
  \label{fig:illustration-dynamic-emoa}
\end{figure}

We adopt the DEMOA introduced in \cite{BGMRT2019}. Note that the focus of this work is on the influence of subsequent decision-making. Therefore, and due to space limitations, we omit most implementation details and present the working principles. For detailed pseudo-code we refer the interested reader to~\cite{BGMRT2019}. Also the implementation is available in a public GitHub repository\footnote{\githubrepo}. We advice the reader to consult Fig.~\ref{fig:illustration-dynamic-emoa} for visual support while reading the following text.

The input for the DEMOA is a problem instance $C = \{1, N\} \cup C^m \cup C^o$, a time resolution $\Delta$, a number of eras $n_t$ and a population size $\mu$. The optimization process starts at time $t = 0$ and the algorithm treats the problem as a sequence of $n_t$ static MOPs (see Section~\ref{sec:probForm}). Note however, that the first era is a special case, since $C^o_{\leq t} = \emptyset$, i.~e., no dynamic requests arrived so far, and there is no possibility to vary the second objective. Hence, in the 1st era, a single-objective Hamiltonian path problem (HPP) on the set $\{1, N\} \cup C^m$ has to be solved. An approximate solution is calculated with the state-of-the-art solver EAX~\cite{nagata_powerful_2013} for the symmetric Travelling-Salesperson-Problem (TSP) after reducing the HPP to a symmetric TSP problem by a sequence of transformations~\cite{jonker:transforming}. Note that in the first era the decision-maker has no choice as there is just a single solution (see era  1 in Fig.~\ref{fig:illustration-dynamic-emoa}). In subsequent eras $j = 2, \ldots, n_t$ time $t = (j-1)\cdot\Delta$ already passed, $C_{\leq t}^o$ is non-empty and as a consequence the problem turns into a true multi-objective problem.
Here, the DEMOA calls a static EMOA whose internals are discussed in the following. The EMOA initializes a multi-set $P$ of $\mu$ candidate solutions. Each candidate solution $x \in P$ is fully described by three vectors of length $N-2$.  A binary vector $x.b = (b_2, ..., b_{N-1})$ indicates which customers are to be visited by the service vehicle (note that the depots 1 and $N$ need to visited in any case and are thus not encoded).\footnote{We want to point out that the current implementation knows the total number $N$ of customers in advance for legacy reasons. However, it only operates on those customers, who asked for service before time $t$. Clearly, it is straight forward to adapt the implementation into a true black-box scenario, where the number of dynamic requests is not known a-priori.} Another vector $x.t$ holds a permutation of $C^m \cup C^o = \{2, \ldots, N-1\}$, i.e., the actual tour where during fitness evaluation only those entries $i$ with $b_i = 1$ are considered. Finally, the vector $x.p = (p_2, \ldots, p_{N-1}) \in [0,1]^{N-2}$ stores per-customer mutation probabilities. If $x.p_i = 0$, the corresponding customer is fixed and not affected by mutation. While in the second era individuals are generated at random (fixing mandatory customers by setting $x.b_i = 1$ and $x.p_i = 0$ for $i \in C^m$), in eras $j \geq 3$ more effort is put into the initialization to transfer as much information from the solution set of the preceding era $j-1$ as possible. The challenge here is that once era $j \geq 3$ starts, the vehicle may already have visited dynamic customers with request times $r_t(i) \leq (j-2)\cdot\Delta$ (this is illustrated by means of example in Fig.~\ref{fig:illustration-dynamic-emoa} last column. Here, bold edges show the fixed, already driven initial tour) given by the decision at the end of the previous era. Fig.~\ref{fig:illustration-dynamic-emoa}). As a consequence, those customers cannot be inactive and hence need to be treated as mandatory customers by the EMOA in all upcoming eras. Moreover, the initial tour, i.e., the part of the tour that has already been driven by the vehicle, needs to be identical for all feasible individuals. Here, the EMOA relies on a sequence of repairing mechanisms.

Given the population $P$ the algorithm continues by adopting a $(\mu + \lambda)$-strategy with NSGA-II~\cite{Deb02} survival selection. Variation is based on feasibility-preserving mutation. Here, each bit $x.b_i$ is flipped independently with probability $p_i$. Subsequently, swap-mutation alters the permutation string $x.t$: with probability $p_{\text{swap}} \in (0, 1)$ a sequence of $\sigma_{\text{swap}}$ exchanges is performed. In addition, every $k$ generations the population is boosted towards shorter tours by applying EAX local-search where the EMOA accounts for the fact that certain nodes have already been visited.
Once the stopping condition has been triggered, e.g., a maximum number of generations has been reached, the solution set is presented to a decision-maker who has to decide on exactly the solution which determines the adaptation of the ongoing vehicle route and which serves as a template for the initialization of the population in the next era.

\section{Computational Experiments}
\label{sec:experiments}
\subsection{Experimental Setup}


In order to gain insights into the decision making process we conducted a  two-stage study. In a first series of experiments we perform a systematic study of decision making strategies. Subsequent experiments focus on a selected sample of decision making strategies on a broader set of instances in order to confirm the lessons learned. 


For the exhaustive experiments we selected three structurally different instances from the pool of instances introduced in ~\cite{MGB2015}: one instance with customer locations spread uniformly at random in the Euclidean plane and two clustered instances with two and three groups of instances respectively. All instances share $|C^m| = 25$ mandatory customers (including depots) and $|C^o| = 75$ dynamic customers with $N = 100$ customers in total. For details on the generation process we refer to~\cite{MGB2015}.
We fixed the number of eras $n_t = 7$ and considered three different ranking based rules for the decision maker in each era. For ranking, the solutions of the approximation set $P_i = \left\{p_1, \ldots, p_{\mu}\right\}$ obtained in era $i$ are sorted in ascending order of tour length and therefore in descending order of the number of unvisited customers. Let $p_{(1)} < p_{(2)} < \ldots < p_{(\mu)}$ denote this order. The \emph{$d$-rank decision maker} ($d \in [0,1]$) then decides for the solution $p_{(k)}$ with $k = \lceil d \cdot \mu\rceil$. Note that small values of $d$ favor solutions with short tours whereas values closer to 1 put a higher emphasis on keeping the number of unvisited dynamic customers low. Our setup considers $d \in D = \{0.25, 0.5, 0.75\}$ in each era\footnote{We do not consider the extremal values $0$ and $1$ for two reasons: (1) to avoid a combinatorial explosion of possible configurations for exhaustive evaluation; (2) extremal decisions are usually unrealistic as they either imply to ignore all optional customers (0) or to accept every optional customer (1).}; we use a tuple notation for sequences of decision maker decisions, termed \emph{decision paths} in the following, e.~g., $(0.25, 0.25, 0.25, 0.25, 0.75, 0.75, 0.75)$ describes the decision path where the DM puts more emphasis on short tours in the first four eras but decides to cover more dynamic requests in the last three eras.
\begin{table}[htbp]
 \centering
 \caption{Dynamic EMOA parameterization.}
 \label{tab:emoa-params}
 \begin{footnotesize}
 \setlength{\tabcolsep}{5pt}
 \begin{tabular}{rl}
    \toprule
    {\bf Parameter}                   & {\bf Setting}\\
    \midrule
    Generations per era               & 65\,000 \\
    $\mu, \lambda$                    & 100 \\
    $p_{\text{swap}}$                 & 0.6 \\
    $\sigma_{\text{swap}}$            & $\nicefrac{N}{10} = 10$ \\
    Local search at generations       & initial, half-time, last \\
    Time limit for local search       & 1s \\
    \bottomrule
 \end{tabular}
  \end{footnotesize}
\end{table}

We run experiments for all $|D|^{n_t} = 2187$ decision paths, i.~e., we cover all kinds of decision scenarios.
At each stage, five independent runs were performed for each of the three instances resulting in a set of $32\,805$ experiments in total.

The results of this \enquote{exhaustive} experimentation served as starting point for subsequent experiments on a broader set of benchmark instances; all 75 instances with $N = 100$ introduced in~\citet{MGB2015}. However, due to combinatorial explosion only a small subset of four decision paths (with different outcomes in the last era) were considered here for an extensive analysis (for details on the selection process, see Section~\ref{sec:results}). For each combination of problem instance and decision path we run the DEMOA 25 times independently in this series of experiments. The parameter configuration of the DEMOA follows the suggestions in \citet{BGMRT2019} and is listed in Table~\ref{tab:emoa-params}.


\begin{figure}[htb]
  \centering
  \includegraphics[width=0.6\textwidth]{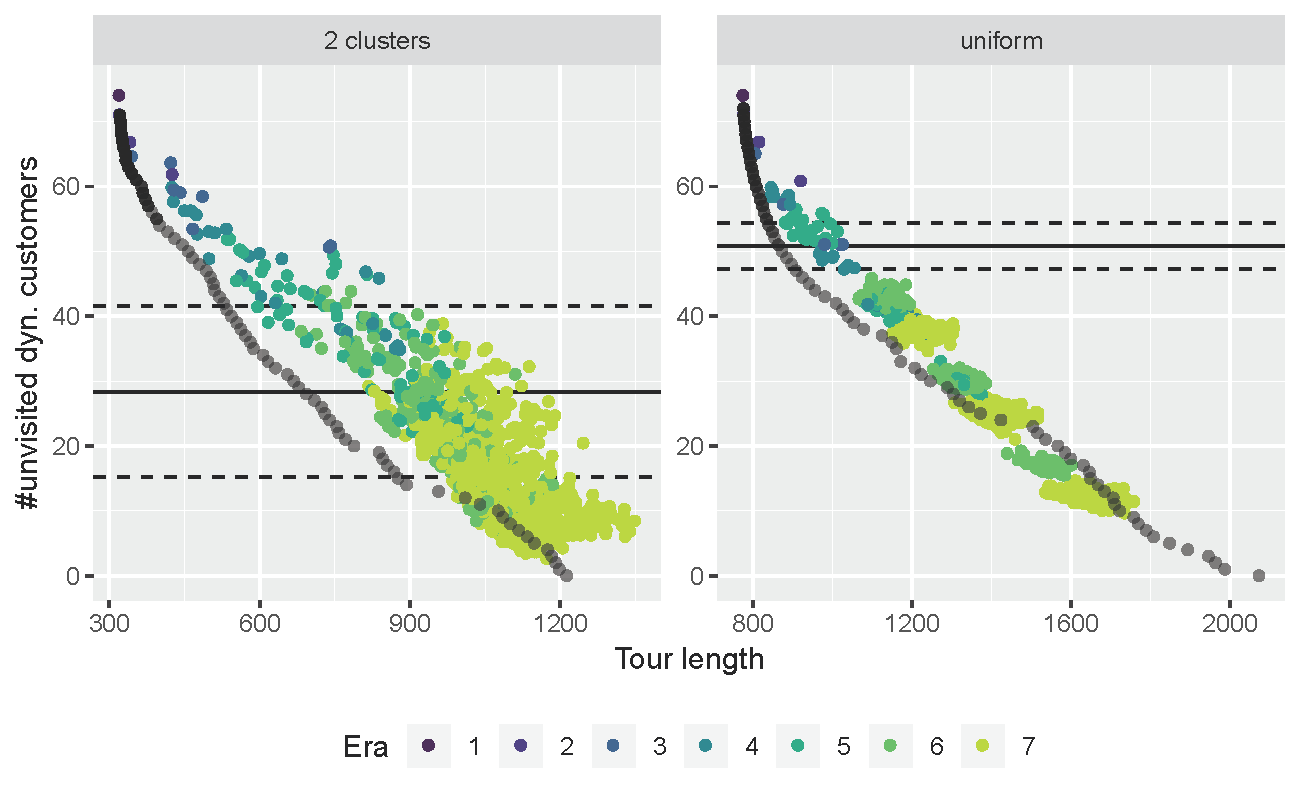}
  \caption{Union of all final decisions made across the complete set of all considered decision maker strategies. Solid and dashed lines indicate the mean upper bound $\pm$ three times the standard deviation of the number of unvisited customers in the last era (based on all decision maker strategies). Small black dots represent the Pareto-front approximation of the clairvoyant EMOA.}
  \label{fig:exhaustive-all-dmpaths}
\end{figure}
\subsection{Results}
\label{sec:results}

Next, we investigate the influence of considered graph topologies as well as the implications of final and intermediary decision making onto the solution development over time. Therefore, we perform a step-wise narrowing of perspective to focus on interesting insights for our considered instances, topologies, and decision strategies in the context of our exhaustive experimental results.

\begin{table}[t]

\caption{\label{tab:upper_bounds}Mean values and standard deviations  of tour lengths and the number of unvisited customers of the solutions in the last era split by the instance type and final decision in the last era.}
\centering
\begin{tabular}{lrrrrr}
\toprule
\multicolumn{1}{c}{\bfseries  } & \multicolumn{1}{c}{\bfseries  } & \multicolumn{2}{c}{\bfseries Tour length} & \multicolumn{2}{c}{\bfseries \#Dyn. customers} \\
\cmidrule(l{2pt}r{2pt}){3-4} \cmidrule(l{2pt}r{2pt}){5-6}
Type & Last decision & Mean & Std & Mean & Std\\
\midrule
 & 0.25 & 997.7 & 59.94 & 22.495 & 5.7486\\

 & 0.50 & 1070.3 & 57.16 & 15.645 & 4.3148\\

\multirow{-3}{*}{\raggedright\arraybackslash 2 clusters} & 0.75 & 1174.9 & 65.14 & 8.105 & 2.4266\\
\cmidrule{1-6}
 & 0.25 & 1227.6 & 28.61 & 37.654 & 0.7662\\

 & 0.50 & 1416.4 & 38.62 & 24.854 & 0.9167\\

\multirow{-3}{*}{\raggedright\arraybackslash uniform} & 0.75 & 1657.1 & 42.32 & 12.069 & 0.8743\\
\bottomrule
\end{tabular}
\end{table}

\subsubsection{General observations}
In Figure~\ref{fig:exhaustive-all-dmpaths} we provide a first overview of the results for all decisions in each era and for all $|D|^{n_t}$decision paths for the uniform instance and the instance with two clusters\footnote{Results for the third instance with three clusters are omitted here since these are very similar to those of the 2-cluster instance.}. Note that in order to compare our results to the clairvoyant EMOA approach -- either visually or by performance metrics -- we transformed the results of all eras to the a-posteriori solution space, which covers the whole potential of arbitrary decision paths.\footnote{We explain this transformation in more in detail here: in the first era, we have zero dynamic requests and consequently zero unvisited dynamic customers. However, in the a-posteriori solution space this solution corresponds to $|C^o|$ unvisited dynamic customers. Therefore, in order to make solutions comparable, a linear transformation of the second objective to the clairvoyant EMOA solution space is required.}

From this high-level perspective, we can identify an interesting property of solution distribution, which not only depends on the applied decision strategies but is strongly related to the considered topology of the instances: For clustered topologies, tour length tends to expose larger variability in later eras, and -- on a first glance -- no matter which combination of decisions is used, while in uniform topologies distinct clusters of solutions represent the different weighting of decision strategies.

\subsubsection{Topology and final decision making influence}
The observed results can be explained by a deeper analysis of the experiments. Here, we focus on two representative topologies consisting of two clusters and a uniformly distributed set of customers, respectively. For both topologies, each instance, and all decision maker strategies, we determine the upper bound of unvisited customers for the last era and compute the mean upper bound as well as standard deviation.
More detailed values for mean and standard deviation of the upper bound of both objectives -- and split according to the final preference -- are available in Table~\ref{tab:upper_bounds}. Each upper bound of unvisited customers is determined by the already visited customers on the traveled partial tour, which is the result of decisions made during previous eras.

For clustered topologies (ref. to Figure~\ref{fig:exhaustive-all-dmpaths}, left) we find a low mean upper bound and large standard deviation, while for the uniform topology the mean upper bound of unvisited customers is rather high with little variance.
Consequently on average, for clustered topologies, the decision maker preference at the onset of the last era allows only little flexibility for final solutions. This leads to the stronger focus on the lower right area of objective space. For uniform instances, the on average larger upper bound of unvisited customers leads to a larger and less flexible range for diverse forming of (intermediary) solutions over all eras and finally to more distinct clustering solutions in objective space.

\begin{figure}[htbp]
  \centering
  \includegraphics[width=0.6\textwidth]{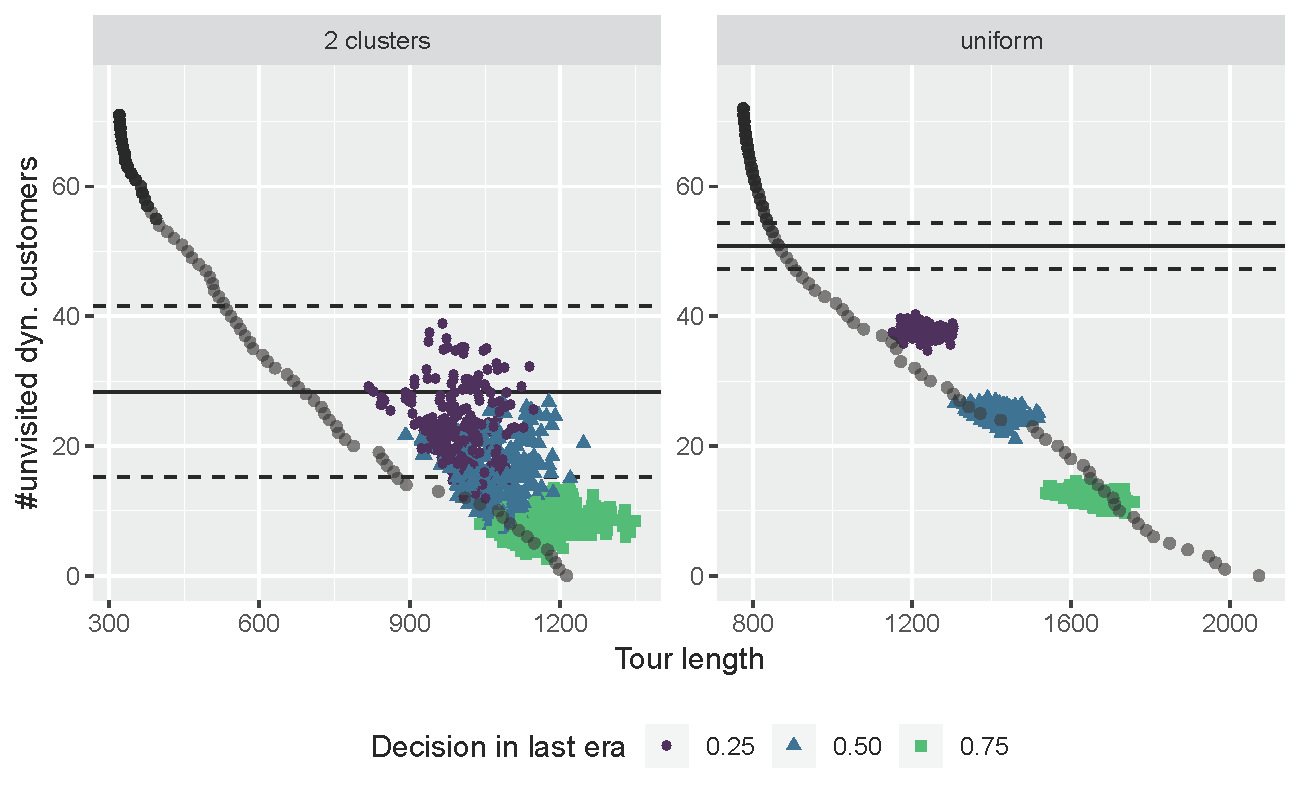}
  \caption{Union of all decisions made in last era colored and shaped by the decision maker preference \underline{in the last era}. Small black dots represent the Pareto-front approximation of the clairvoyant EMOA.}
  \label{fig:dm-in-last-era}
\end{figure}

The coloring of final solutions with respect to the final decision maker preference in Figure~\ref{fig:dm-in-last-era} provides additional insights into the partitioning of the exhaustively generated solutions. In fact, we find that in both cases the last decision preference has significant influence on the solution position. For the uniform instances, however, the preferences are more distinct due to more certain planning flexibility in the final era.


From Figure~\ref{fig:dm-in-last-era} and Table~\ref{tab:upper_bounds} we also conclude that solutions for uniform topologies expose less variance in quality and converge closer to the a-posteriori solution fronts determined by application of the clairvoyant EMOA. As a consequence, more solutions of the dynamic approach individually outperform solutions on the a-posteriori front.\footnote{The a-posteriori front used here was achieved as non-dominated set of the union of ten clairvoyant EMOA runs.} This effect is rooted in decreasing complexity of the tour planning component of the bi-objective problem under the successive dynamic decision making~\citep{BGMRT2019}. Due to decision making over time eras, partial tours are already completed such that the combinatorial decision space shrinks to the still available customers leaving the tour planning problem with less degrees of freedom. Clearly, this observation holds for clustered instances, too. Here, the mean upper bound is even lower. However, in the clustered setting the service vehicle might need to travel back and forth between clusters in order to fulfill the decision maker preferences which oftentimes might lead to enlarged tour length in particular in late eras and a high preference on the second objective. We will catch up on this important aspect later on.

\subsubsection{Intermediate decision making}
While so far we analyzed the final decision maker preference for some instances, the following stage of investigation is focused on the influence of intermediary decisions. A subset of decision paths, which led to the non-dominated as well as completely dominated solutions in the last era over all considered topologies is selected.
For this selection, we detail the effects of decision steps that yielded very good and very bad results, compared on final solutions. For the following discussion, we investigate the results up to specific solution phenotypes, i.~e., the development of specific tours over time. We present detailed results for two exemplary but representative out of 75 instances.

\begin{figure*}[htbp]
  \centering
  \includegraphics[width=0.49\textwidth]{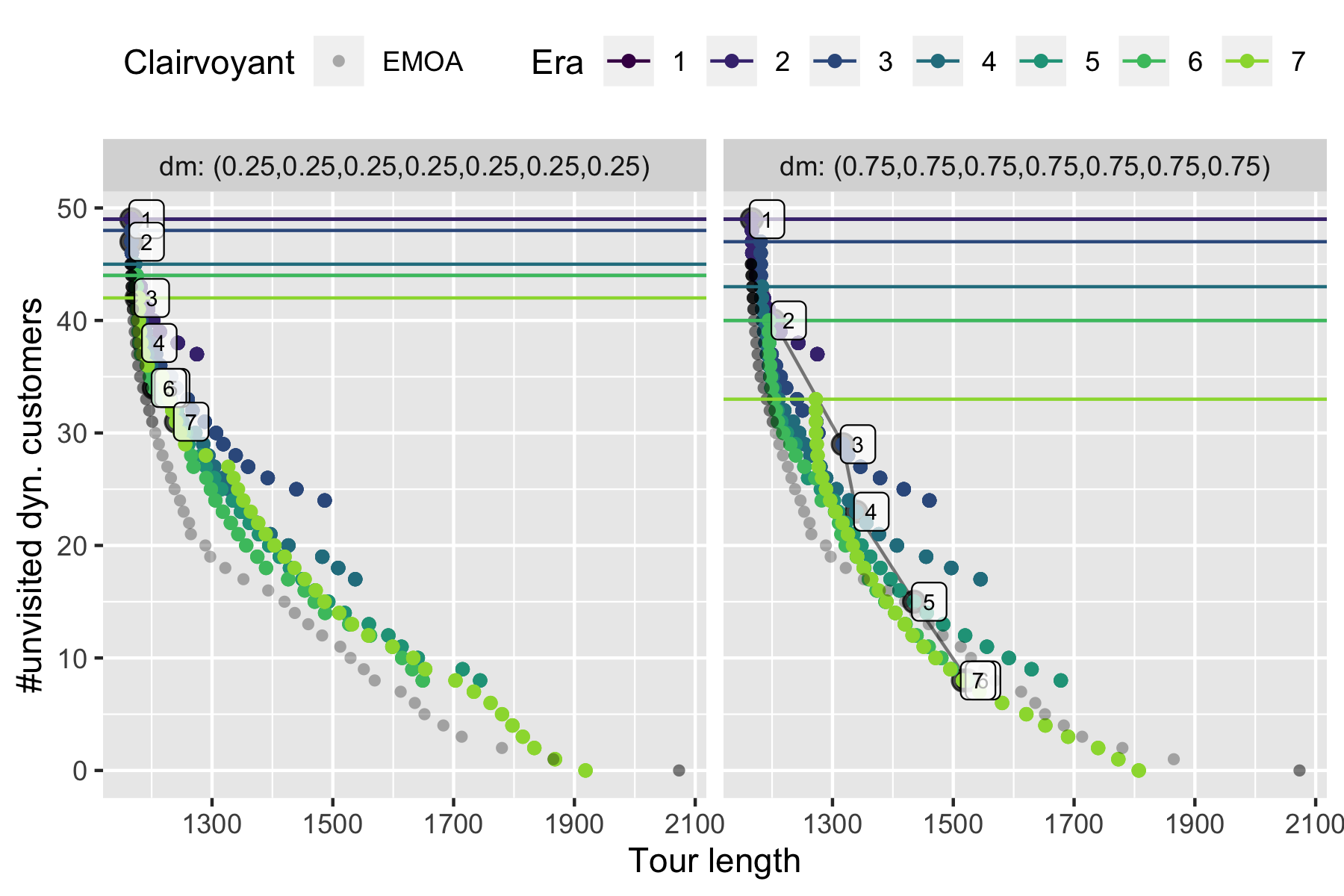}
  \hfill
  \includegraphics[width=0.49\textwidth]{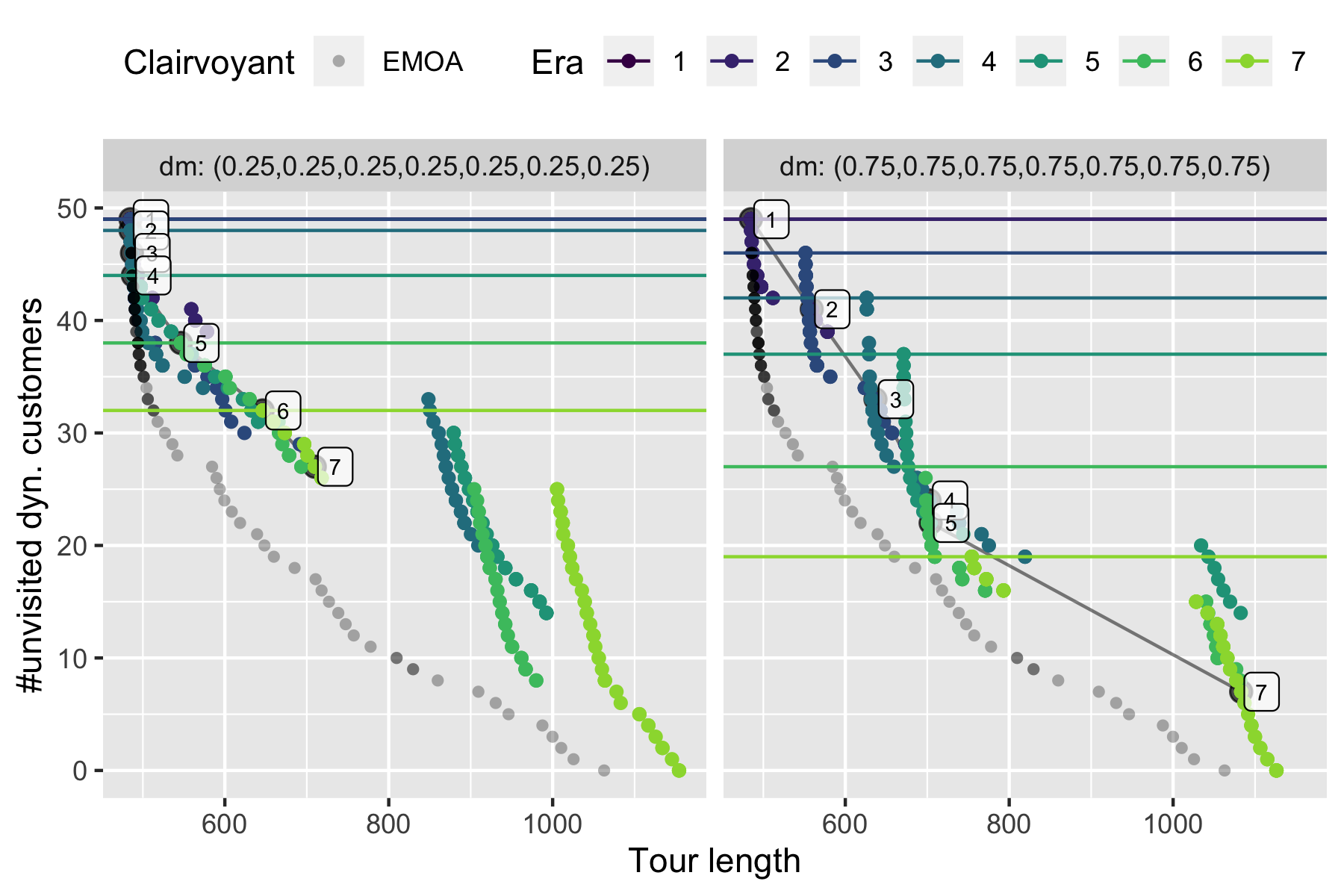}

  \includegraphics[width=0.49\textwidth]{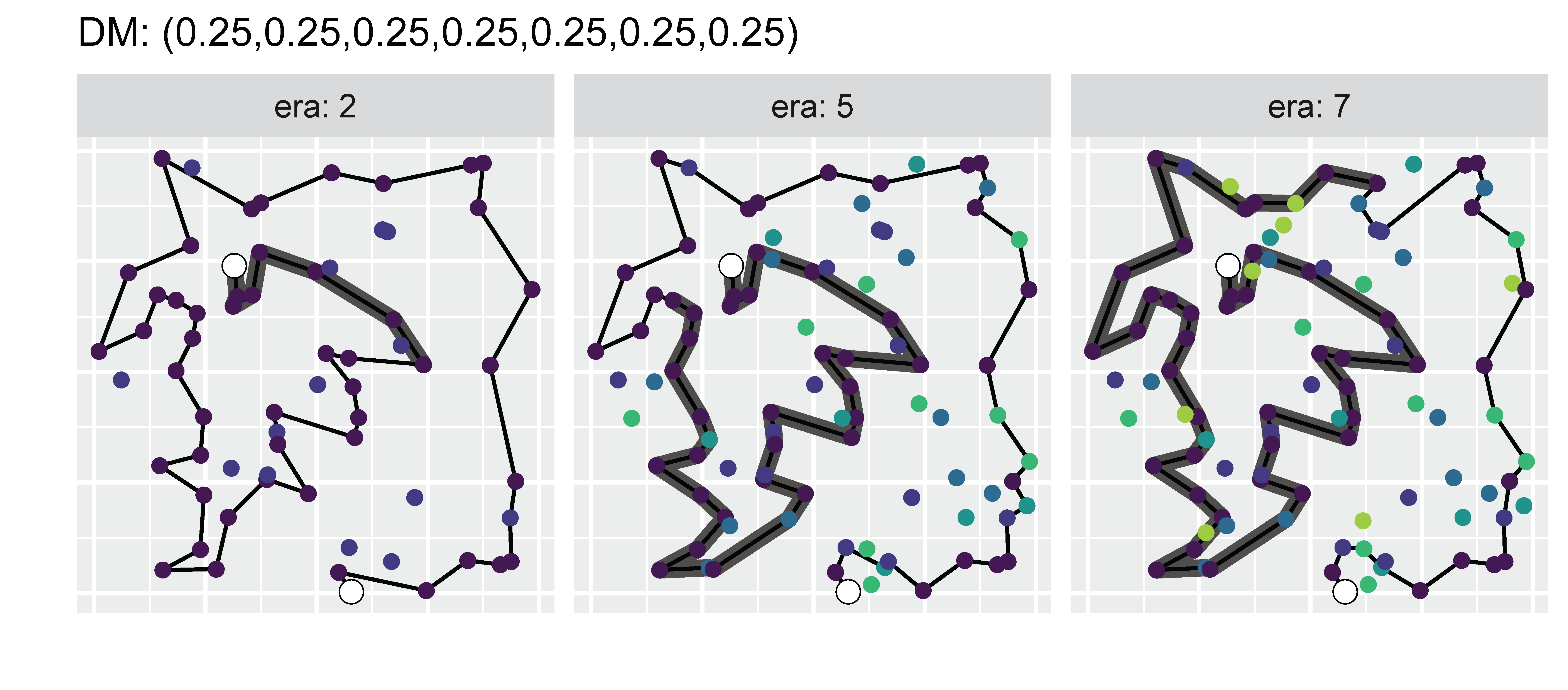}
  \hfill
  \includegraphics[width=0.49\textwidth]{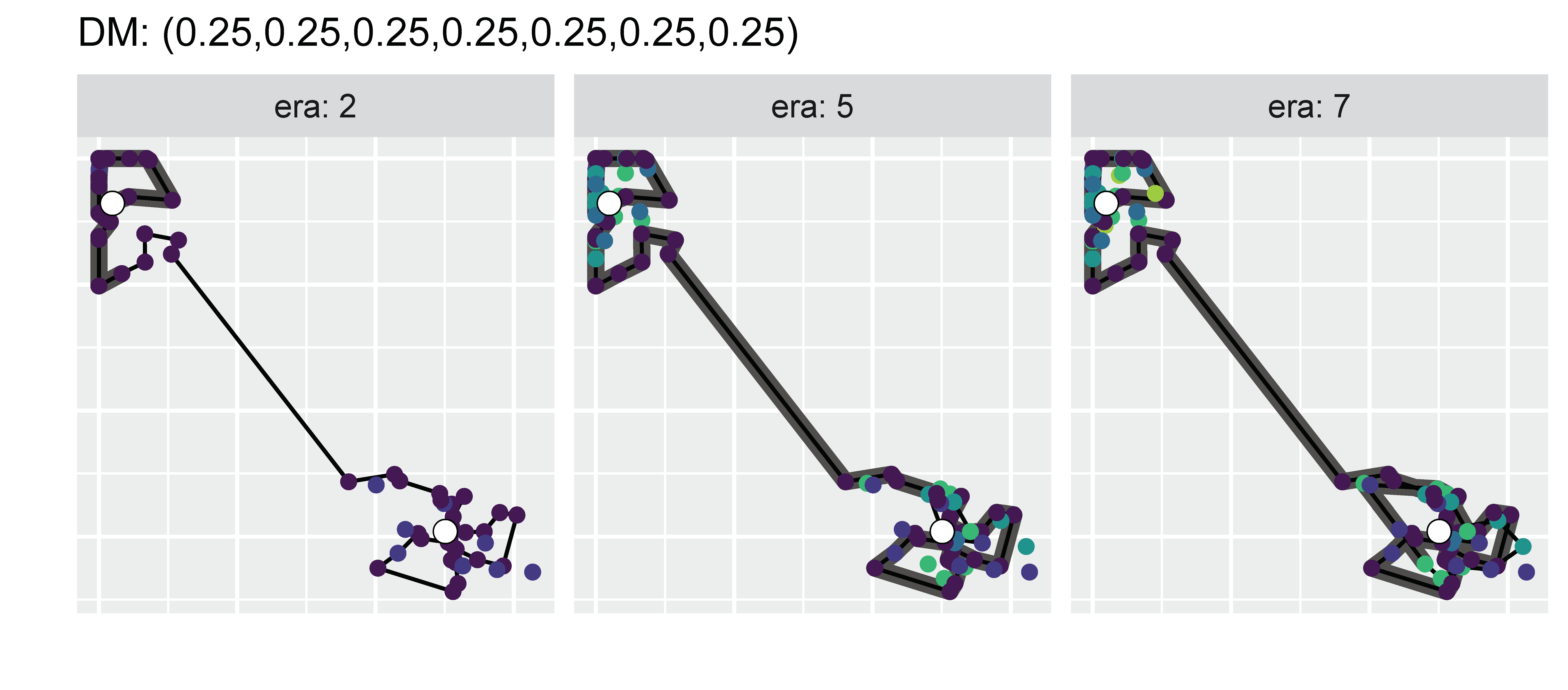}

  \includegraphics[width=0.49\textwidth]{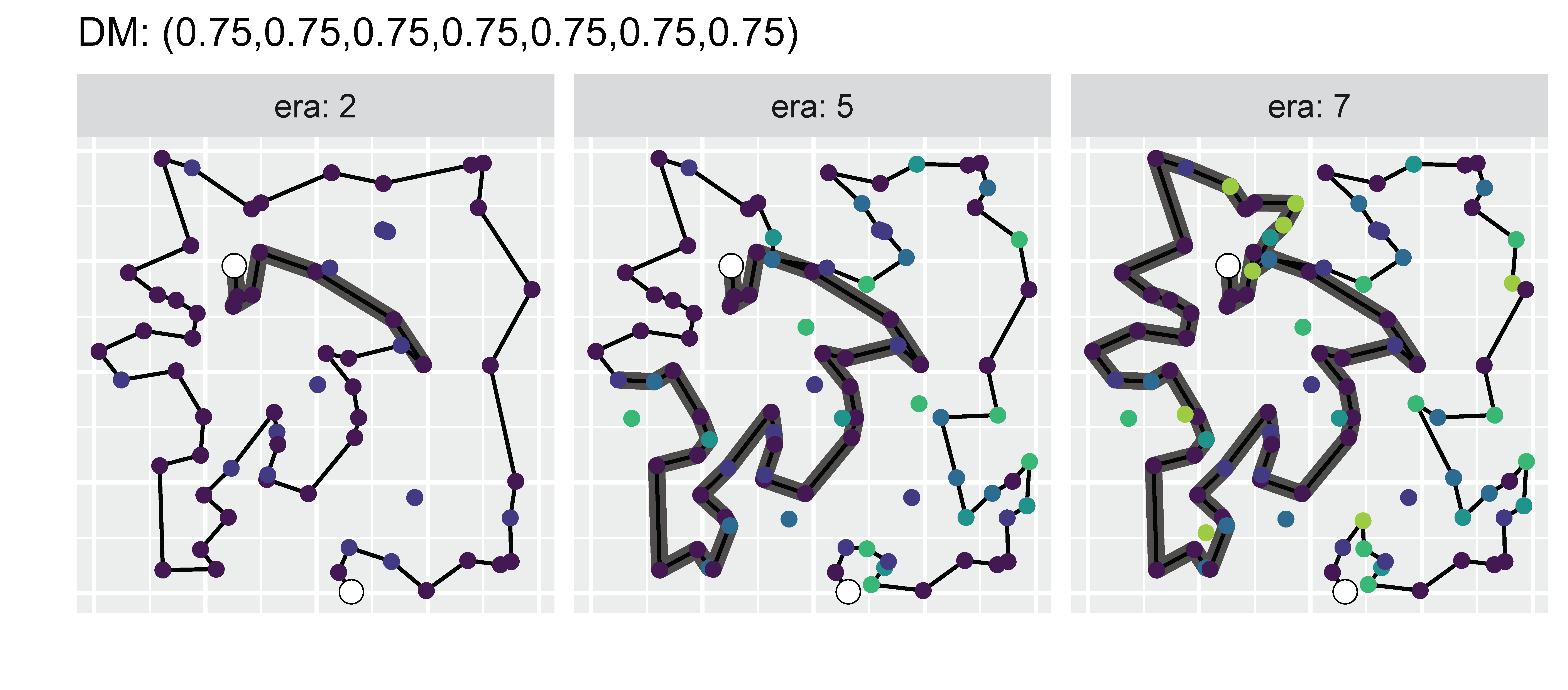}
  \hfill
  \includegraphics[width=0.49\textwidth]{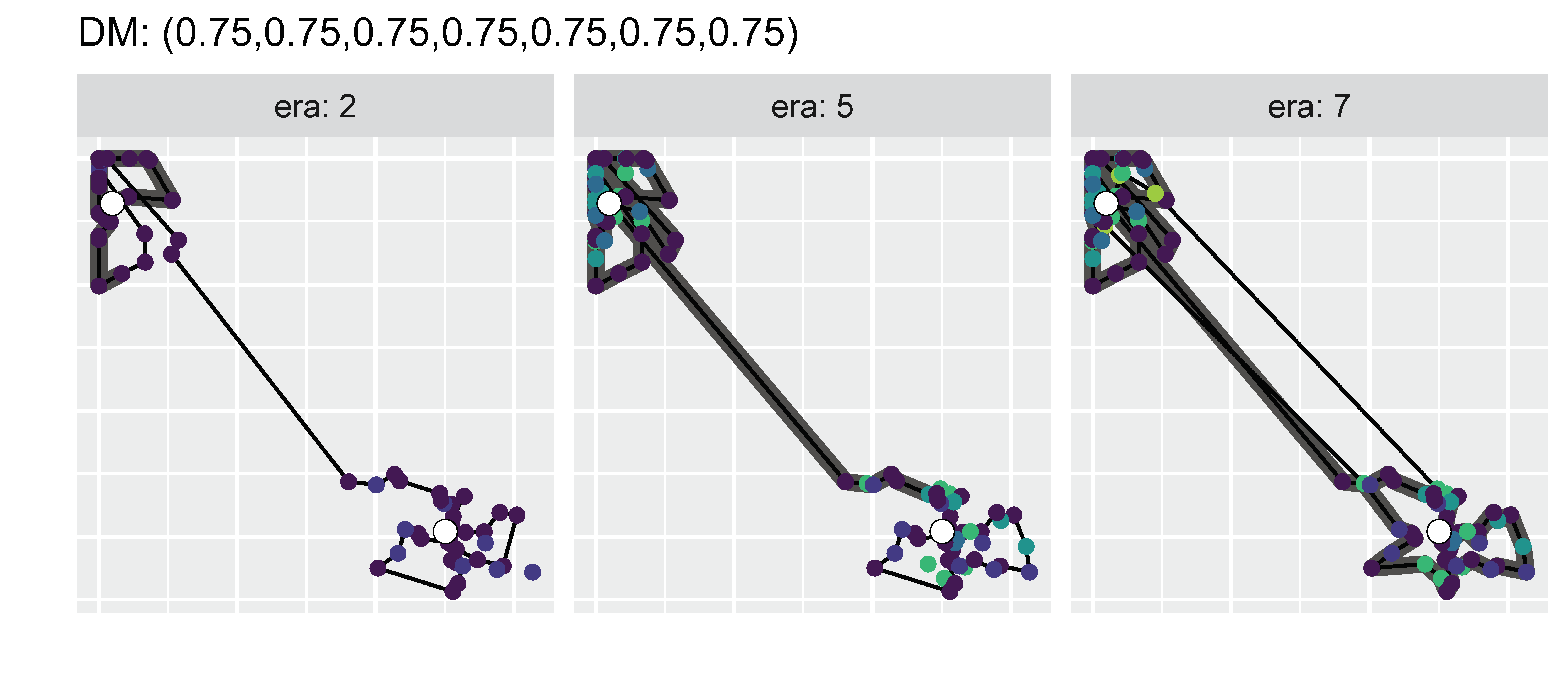}

  \caption{Pareto-front approximations for eras 2, 5, 7 and two selected decision maker strategies (top row plots) for an exemplary uniform instance (left side) and an instance with two clusters (right side) with 50 dynamic customers. Colored horizontal lines in the top plots show the upper bound for unvisited customers in the respective era. Numeric labels $i \in \{1, \ldots, 7 = n_t\}$ indicate the decision made by the DM in the corresponding era. In the rows below phenotypes of the decisions are plotted for the respective era aus DM stratigies.}
  \label{fig:eraplots_and_tours}
\end{figure*}

Figure~\ref{fig:eraplots_and_tours} provides detailed insight into the development of solutions under different DM strategies. In order to visualize the effect of the permutation of decisions inside a strategy (and also due to space limitations), we restrict ourselves to one uniform topology and one topology with two-clusters again and show results of single representative runs. At the top of the figure we show the non-dominated solutions of all eras of these topologies regarding four strategies that follow (a) only $0.25$ preferences, (b) first four times $0.25$ and then three times $0.75$ preferences, (c) the inverse strategy to (b), and (d) only $0.75$ decisions. Below the non-dominated solutions, we visualize the development of exemplary tours of the solutions. We omit era~1, where the vehicle has not traveled yet and also omit some intermediate tours to show a second example tour. For each tour, the decision path via intermediate solutions is included into the respective top figure and annotated with the era number.

For the clustered topology, we find a strategy-sensitive behavior that is related to when (in which era) preferences are used. The overall observation is that preferences, which do not put a strong focus on minimizing the number of unvisited customers (represented by a sequence of only $0.25$ preference) lead to rather short tours (according to the second objective). In these tours, the vehicle transfers to the other cluster only once. Introducing a strong preference for visiting all customers (represented by a sequence of only $0.75$ preferences) forces the vehicle to transfer multiple times between clusters, see Figure~\ref{fig:eraplots_and_tours} bottom right plot. This behavioral changes are also observable for planned tours, when preferences mix, e.g., when a strong preference for visiting many customers is only present at the beginning or the end of the strategy sequence. In many sequences with changing decision preferences (not shown here as figure), we observe that planned transfers  in early eras vanish in following eras (due to $0.25$ preferences later). From this behavior we conclude, that intermediary preference ordering can have decisive influence on the solution generation process for clustered instances. With respect to minimization of unvisited customers and dynamic appearance of customers, decision maker preferences have different degrees of greediness: a $0.25$ preference is far less greedy than a $0.75$ strategy and often allows the vehicle during tour planning to remain in the current cluster, as far fewer customers need to be served.

Introducing a more greedy strategy often forces a vehicle transfer to serve the preferred amount of customers. The observations however show, that flexible re-planning is still possible as long as the partial tour has not been realized.


Considering the exemplary but representative results from Figure~\ref{fig:eraplots_and_tours}, we can conclude, that strategy preferences are less important compared to the clustered case. As mandatory and dynamic customers are uniformly distributed, planned tours do not need to be changed extensively in order to visit or ignore a customer. When we switch preferences from less greedy to more greedy, customers \enquote{on the way} can be included. The same way, dynamic customers can be excluded again from a planned tour, often without significant changes in the overall tour length.

The described effects for clustered and uniform topologies can also be observed in the resulting non-dominated fronts for the eras. For clustered topologies, the front usually exposes a gap, which corresponds to the additional traveled distance in size. It appears, that solutions cannot be realized without transferring the vehicle multiple times. Such case usually does not happen for uniform topologies such that the approximated efficient front does not expose a gap.

\subsubsection{Performance measurement}

\begin{figure*}[htb]
  \centering
  \includegraphics[width=0.95\textwidth]{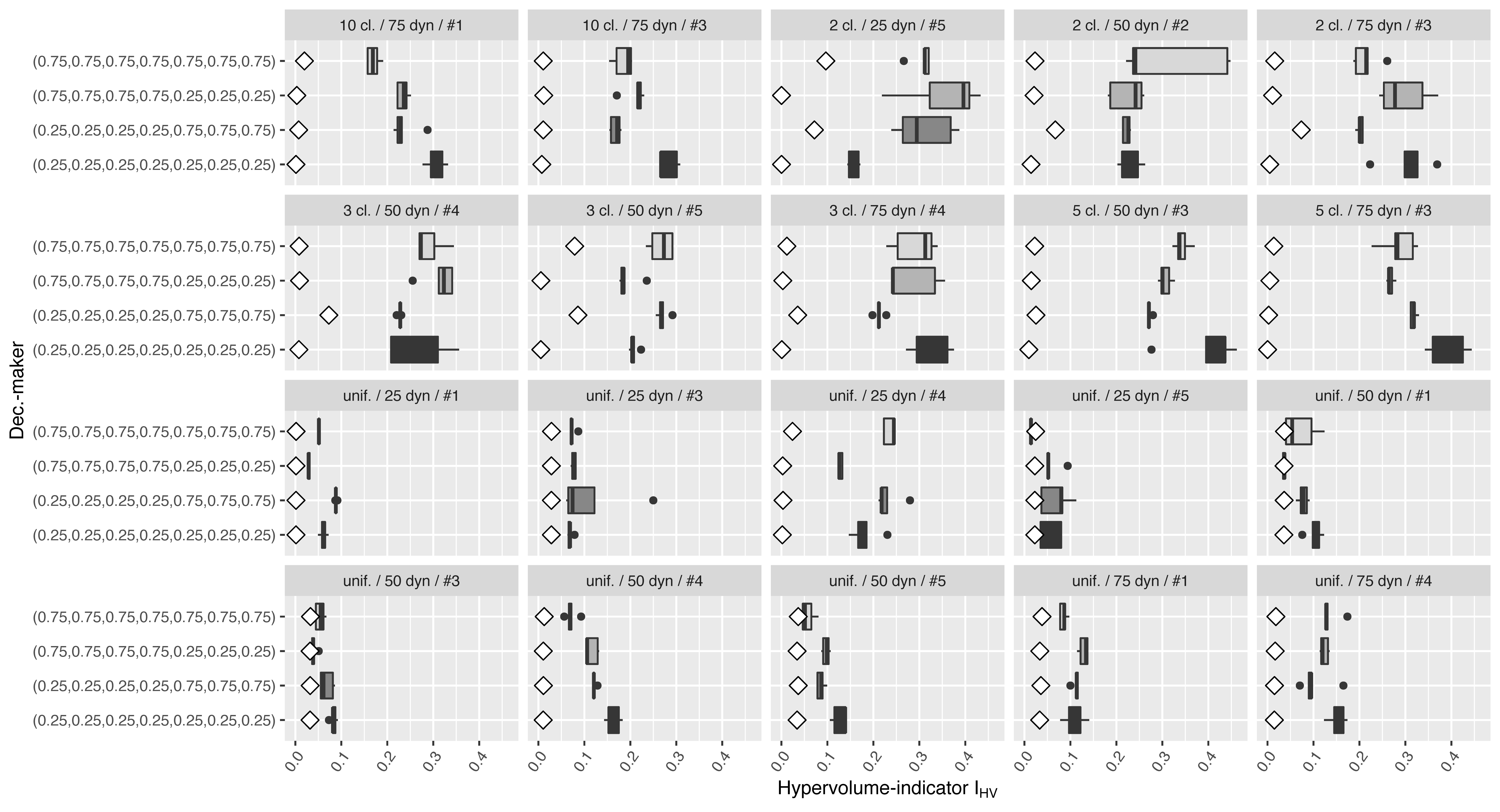}
  \caption{Distribution of hypervolume-indicator $I_{\text{HV}}$ (lower is better) for 20 out of 75 randomly sampled instances. The white diamonds indicate the HV-indicator for the union of 10 runs of the clairvoyant EMOA, i.~e., the baseline.}
  \label{fig:performance-hv}
\end{figure*}
In order to support our observations from the previous paragraphs we continue with indicator-based performance assessment of the DEMOA in comparison to the approximation sets calculated by the clairvoyant EMOA. We aim to quantify the quality of the overall final approximation set in the last era. We use the hypervolume indicator $I_{\text{HV}}(P, R)$~\citep{ZDT00} to measure the space enclosed by a reference set $R$ (non-dominated set of the union of clairvoyant EMOA approximation and all front approximations for the problem instance obtained by the DEMOA) and the DEMOA approximation $P$.
We restrict our analysis to the DEMOA approximation sets of the final era only. We take account for the upper bound that restricts the possible number of unvisited dynamic customers in the last era as follows: only solutions of the clairvoyant EMOA whose second objective is lower or equal to the maximum upper bound in the last era for each instance over all 25 performed runs are taken into consideration. We want to stress that this comparison -- and the one in the next paragraph -- is obviously highly unfair, i.~e., (1) the clairvoyant EMOA has a clear advantage over the dynamic approach due to its a-priori knowledge of request times and (2) the solutions of the clairvoyant might not even be feasible anymore in the last era. Hence, we do not expect the DEMOA to beat the clairvoyant EMOA by any means. Instead, our goal is to learn how close we can approach the clairvoyant solutions with the dynamic approach.
Figure~\ref{fig:performance-hv} shows the distributions of the HV-indicator split by instance and the four DM-strategies discussed before. We show results for a random sample of 10 uniform and 10 clustered instances. The plots confirm our previous observations: in the case of customers distributed uniformly at random in the Euclidean plane the final approximation sets are close to the reference set. In contrast, for clustered topologies the situation is different. Here, as the vehicle possibly needs to transfer between clusters multiple times, the oracle-perspective of the clairvoyant EMOA is much more advantageous and has a much larger impact. In other words, the HV-indicator is less close to the one of the clairvoyant EMOA.

\section{Conclusions}
\label{sec:conclusion}



%
%

For bi-objective vehicle routing, problem dynamics have to be efficiently addressed while suitable decision maker strategies accounting for the trade-off of minimizing overall tour length and maximizing the number of served customers are required simultaneously. We  build upon previous work which provides a sophisticated dynamic EMOA hybridized with local search and specifically investigate the influence of respective decision maker preferences and strategies.

As vehicle tours for a given problem instance evolve over the focused time horizon, decision maker preferences regarding both objective functions may change in the course of the day. We assume that the decisions for possibly altering a predefined tour based on new customer requests have to be made at predefined time intervals which of course subsequently impacts optimization algorithm behavior and thus also influences solution selection decisions which have to be made at later stages.

In systematic experiments, we investigated the influence of decision paths, i.e. sequences of (possibly different) decision maker preferences and solution selections. We present a decision support system enhanced by informative figures visualizing the vehicle tour over time and the characteristics of the candidate trade-off solutions at the points of required decisions.

We confirm the reasonable suspicion that decision making is sensitive to the underlying problem topology. For clustered topologies, intermediate decisions should be considered carefully, as too greedy approaches can lead to multiple vehicle transfers between clusters and massively deteriorate solution quality. For uniform instances, sensitivity is low and the last decision for the optimal trade-off solution is of major importance for final tour quality. Consequently, it is important for the decision maker to estimate the customer location topology for adjusting the greediness of decision making.

Future work directions are manifold with the most promising being listed below:
\begin{itemize}
    \item The problem can be extended to a more realistic scenario which includes multiple vehicles or vehicle loading and unloading during service. Thereby, properties and challenges of the traveling thief problem~\citep{Bonyadi2013,Blank2017} are included into consideration.
    \item We see much room for algorithmic improvements. The insights gained in this paper suggest that biased mutation (e.~g., activating customers in the current cluster with higher probability) may have beneficial effects on the solution quality of the DEMOA. Furthermore, utilizing probabilistic models to predict upcoming customer requests can be leveraged to achieve more thoughtful algorithmic tour planing.
    \item Last but not least the major goal is to refine the presented approach in terms of providing tool-support for informative interactive decision making in this highly dynamic environment. Moreover, we will include predefined agent-based decision maker paths into the algorithm which adapt to problem topology characteristics via automatically extracting problem features and which can be adjusted if needed.
\end{itemize}

\bibliographystyle{unsrt}
\bibliography{references}  

\end{document}